\pdfoutput=1

\documentclass[11pt]{article}

\usepackage{acl}

\usepackage{times}
\usepackage{latexsym}

\usepackage[T1]{fontenc}

\usepackage[utf8]{inputenc}

\usepackage{microtype}

\usepackage{graphicx}
\usepackage{amsfonts}
\usepackage{amsmath}
\usepackage{amssymb}
\usepackage{amsthm}
\usepackage{cite}
\usepackage{bm}
\usepackage{xspace}
\usepackage{pifont}
\usepackage{algorithm}
\usepackage{algpseudocode}
\usepackage{booktabs}
\usepackage{multirow}
\usepackage{multicol}
\usepackage{listings}
\usepackage{verbatim}
\usepackage{caption}
\usepackage{todonotes}
\usepackage{mathtools}
\def\eqref#1{equation~\ref{#1}}

\def\1{\bm{1}}

\def\vs{{\bm{s}}}

\def\vx{{\bm{x}}}

\DeclareMathAlphabet{\mathsfit}{\encodingdefault}{\sfdefault}{m}{sl}
\SetMathAlphabet{\mathsfit}{bold}{\encodingdefault}{\sfdefault}{bx}{n}

\DeclareMathOperator*{\argmax}{arg\,max}

\newcommand{\eg}{\emph{e.g.}}
\newcommand{\ie}{\emph{i.e.}}

\newcommand{\ours}[0]{\textsc{SelfDenoise}\xspace}
\newtheorem{theorem}{Theorem}

\title{Certified Robustness for Large Language Models with Self-Denoising}

\author{Zhen Zhang$^1$, Guanhua Zhang$^1$, Bairu Hou$^1$, Wenqi Fan$^2$, Qing Li$^2$, \\ \textbf{Sijia Liu}$^{3,4}$, \textbf{Yang Zhang}$^4$, \textbf{Shiyu Chang}$^1$\\
$^1$UC Santa Barbara $^2$The Hong Kong Polytechnic University\\$^3$Michigan State University $^4$MIT-IBM Watson AI Lab\\
\{zhen\_zhang, guanhua, bairu, chang87\}@ucsb.edu, wenqifan@polyu.edu.hk, \\ csqli@comp.polyu.edu.hk, liusiji5@msu.edu, yang.zhang2@ibm.com
}

\begin{document}
\maketitle

\begin{abstract}
Although large language models (LLMs) have achieved great success in vast real-world applications, their vulnerabilities towards noisy inputs have significantly limited their uses, especially in high-stake environments. 
In these contexts, it is crucial to ensure that every prediction made by large language models is stable, \emph{i.e.}, LLM predictions should be consistent given minor differences in the input. 
This largely falls into the study of certified robust LLMs, \emph{i.e.}, all predictions of LLM are certified to be correct in a local region around the input. 
Randomized smoothing has demonstrated great potential in certifying the robustness and prediction stability of LLMs.
However, randomized smoothing requires adding noise to the input before model prediction, and its certification performance depends largely on the model's performance on corrupted data.
As a result, its direct application to LLMs remains challenging and often results in a small certification radius.
To address this issue, we take advantage of the multitasking nature of LLMs and propose to denoise the corrupted inputs with LLMs in a self-denoising manner.  
Different from previous works like denoised smoothing, which requires training a separate model to robustify LLM, our method enjoys far better efficiency and flexibility.
Our experiment results show that our method outperforms the existing certification methods under both certified robustness and empirical robustness.
The codes are available at \url{https://github.com/UCSB-NLP-Chang/SelfDenoise}.
\end{abstract}

\section{Introduction}
\label{sec:introduction}
\begin{figure}[t]
    \centering
    \includegraphics[width=0.45\textwidth]{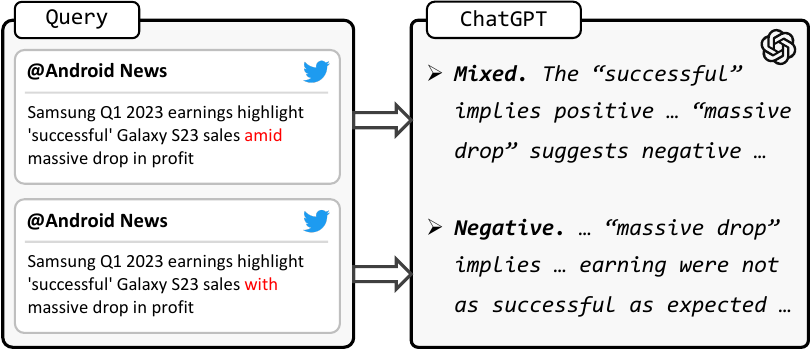}
    \caption{ \small
    Prompting LLMs for Tweet sentiment analysis. The state-of-the-art ChatGPT language model shows vulnerabilities to minor changes in the input.}
    \label{fig:intro-example}
    \vspace{-5mm}
\end{figure}

Large language models have shown exceptional performances in vast applications~\citep{touvron2023llama,Wu2023BloombergGPTAL,Taylor2022GalacticaAL,li2023empowering,Yang2022ALL,Nijkamp2022CodeGenAO}, even outperforming humans over multiple benchmarks~\citep{chowdhery2022palm}.
However, unlike human intelligence, LLMs are vulnerable to noises and perturbations on the input which does not change the semantic meaning.
For example, as shown in Figure~\ref{fig:intro-example}, with minor changes in the input, the state-of-the-art ChatGPT model gives opposite predictions.
Such vulnerability has impeded LLMs from being used in high-stake environments, like financial and medical applications, where prediction stability and reliability are crucial.
To address the problem, it largely falls into the study of certified robustness~\citep{Cohen2019CertifiedAR}, which ensures that all predictions made by the model are correct within a local region around the input.

The enormous model size and limited access to parameters of LLMs have brought great obstacles to most certification techniques~\citep{Shi2020RobustnessVF}.
As a result, as far as we know, the only potential way to provide a certified robustness guarantee for LLMs is randomized smoothing, which converts the original LLM into a smoothed model~\citep{zeng2021certified}.
However, the certification performances by directly applying randomized smoothing in LLMs are still far from satisfactory.
The underlying reason is that, randomized smoothing requires adding noise to the input before model prediction, and its certification performance depends largely on the LLM's performance on corrupted data.
Several previous works alleviate the problem by fine-tuning the model with noisy inputs for a certain task, while this is infeasible for LLMs due to the partial access to parameters and the huge computational costs for fine-tuning.

To address this issue, in this paper, we propose \ours, a self-denoising LLM certification framework based on randomized smoothing.
The proposed approach first generates multiple perturbed inputs by randomly masking words in the original input. 
Different from vanilla randomized smoothing which directly feeds these perturbed inputs to the model, we additionally denoise these perturbed inputs by using the LLM itself as a denoiser.
Specifically, the perturbed inputs are fed to the LLM, and the LLM is asked to complete the sentences by filling in the masked parts. 
The resulting sentences are then forwarded to LLM for performing certain downstream tasks such as sentiment analysis. 
Such a denoising mechanism is inspired by denoised smoothing~\citep{Salman2020DenoisedSA}, where a separate model is trained to robustify the base model.
Extensive experiments are conducted on two datasets using state-of-the-art LLM, Alpaca, and the results show our superiority over baselines on both certified and empirical robustness.
\vspace{-2mm}
\section{Related Work}
\label{sec:related_work}
Certifying the robustness of neural networks is challenging due to the non-convexity and the growing size of neural networks. The mainstream of existing works can be divided into two main categories: \ding{192} linearization-based verification that is often based on the branch and bound (BaB) technique~\citep{
zhang2019efficient, singh2019abstract, gehr2018ai2, bonaert2021fast, pmlr-v80-mirman18b,Jia2019CertifiedRT,Huang2019AchievingVR}.
\ding{193} certification with randomized smoothing~\citep{Cohen2019CertifiedAR, Salman2020DenoisedSA,Levine2019RobustnessCF,Zhao2022CertifiedRA,zeng2021certified,Ye2020SAFERAS}. 
Linearization-based method recursively splits the original verification problem into subdomains (\emph{e.g.}, splitting a ReLU activation into positive/negative linear regions by adding split constraints). 
Then each sub-domain is verified with specialized incomplete verifiers. 
With the enormous model size and non-linear operations (\emph{e.g.}, self-attention), it is very challenging to verify LLMs.
The discrete nature of text data makes certification even more difficult as it poses extra challenges on optimization. 
Due to the difficulty of applying linearization-based methods on LLMs, we focus on randomized smoothing-based methods. 

Several existing works have adopted  randomized smoothing in the NLP domain, where noises are added to the input by uniformly sampling some positions in the input and then mask them~\citep{zeng2021certified} or replace them with their synonyms~\citep{Ye2020SAFERAS, wang2021certified,Zhao2022CertifiedRA}. 
Among them, the methods that replace selected tokens with synonyms~(\eg, \textsc{Safer}, \citet{Ye2020SAFERAS}) introduce additional assumptions on the perturbations. 
However, in realistic scenarios, we do not have full knowledge about the potential perturbations, making these methods less practical. 
Therefore, in this paper, we add noises by masking the selected tokens, \emph{i.e.}, replacing them with \emph{[MASK]}. 
Besides, the certification performance of randomized smoothing depends largely on the model's performance on masked inputs.
Existing methods fine-tune the base model~\citep{zeng2021certified,Zhao2022CertifiedRA} or train an additional denoiser ~\citep{Salman2019ACR}, which requires access to the LLM parameter and huge computational costs.
In contrast, we propose a self-denoising framework where LLM itself is used as the denoiser for free.
\vspace{-2mm}
\section{Preliminaries and Notation}
\label{sec:preliminary}
For a certain task, we denote $\vx=[x_1, x_2, \ldots, x_L]$ as the input to the LLM $f(\cdot)$, where $x_i$ is the $i$-th token, and use $y \in \mathcal{Y}$ as the ground truth output.

\paragraph{Certified robustness}
The model $f(\cdot)$ is certified robust if it satisfies following condition for any $\vx$,
\vspace{-1mm}
\begin{equation}
\small
    f(\vx') = y\text{,}~~||\vx' - \vx||_0 \leq dL\,\text{,}
\end{equation}

\vspace{-1.5mm}
\noindent where we use $||\vx' - \vx||_0$ to denote the Hamming distance, \ie, $\sum_{i=1}^{L}\mathbb{I}(x'_i \neq x_i)$ with $\mathbb{I}(\cdot)$ as the indicator function, and $d$ refers to perturbation scale.
A certified robust LLM is expected to generate the correct output $y$, given at max $d$ percentage word perturbation on the input.
Our definition for robustness differs from previous works~\citep{Ye2020SAFERAS} in that we do not assume a synonym candidate list for word replacement in $\vx'$, \ie, each position could be replaced to any word, to mimic the vast kinds of noisy inputs in real-world applications.

\paragraph{Randomized smoothing}
Randomized smoothing robustify the original LLM $f(\cdot)$ by turning it into a smoothed model $g(\cdot)$, which returns the most likely output predicted by $f(\cdot)$, \ie,
\begin{align}
\footnotesize
    g(\vx) = \argmax_{c\in\mathcal{Y}} \underbrace{
    P_{\vs \sim \mathcal{U}(L, m)}(f(\mathcal{M}(\vx, \vs))=c)
    }_{p_c(\vx)}\text{,}
\end{align}

\vspace{-1mm}
\noindent where we introduce $\vs$ as a mask position selector, sampled from a uniform distribution $\mathcal{U}(L,m)$ over all possible sets of $mL$ unique indices of $\{1,\ldots,L\}$. 
$\mathcal{M}$ refers to the masking operation, which masks the corresponding $m$ percent words indicated by $\vs$ with \emph{[MASK]}.
$p_c(\vx)$ refers to the probability that $f$ returns class $c$ after random masking.
The smoothed classifier predictions are certified to be consistent with input perturbations,
\begin{theorem}
    For any $\vx$, $\vx'$, $||\vx-\vx'||_0 \leq dL$, if
    \vspace{-2mm}
    \begin{align}
        \underline{p_c(\vx)} - \beta \Delta > 0.5\,\text{,}
        \label{eq:condition}
    \end{align}
    \vspace{-6mm}
    
    \noindent then with probability at least $(1-\alpha)$, $g(\vx')=c$.
\end{theorem}
\noindent where $\underline{p_c(\vx)}$ refers to a lower bound on $p_c(\vx)$.
$\beta$ is set to $1$ in \citet{Levine2019RobustnessCF} and approximated with $p_c(\vx)$ in \citet{Zeng2021CertifiedRT}.
$\Delta=1-\binom{\tiny L-dL}{\tiny L-mL} / \binom{\tiny L}{\tiny L-mL}$ is determined by the input length $L$, masked word percentage $m$ and perturbation scale $d$.
We refer the readers to \citet{Zeng2021CertifiedRT,Cohen2019CertifiedAR} for detailed calculation of $\underline{p_c(\vx)}$, $\beta$ and $\Delta$, and the related proof.

In practice, for a certain $\vx$ and scale $d$, one could try different values of masked word percentage $m$ to calculate the corresponding $\underline{p_c(\vx)}$, $\Delta$ and $\beta$. 
The model $g(\cdot)$ is certified to be robust on $\vx$ with scale $d$  if the probability that $f$ returns ground truth label $\underline{p_y(x)} - \beta\Delta>0.5$, following Equation~\ref{eq:condition}.
We then use $r=\max_{(\underline{p_y(x)}-\beta\Delta>0.5)} d$ as the certification radius on $\vx$, \ie, perturbations with at most $d$ percent words cannot alter model prediction.

\vspace{-2mm}
\section{Methodology}
\label{sec:methodology}
The performance of randomized smoothing largely depends 
on  $p_y(\vx)$, which is determined by the performances of the base model $f(\cdot)$ on the masked inputs $\mathcal{M}(\vx, \vs)$.
However, naively applying the randomized smoothing on the base LLM could give a small certification radius as the LLMs are not trained to be robust to random masks on the inputs for downstream tasks.
As discussed, many previous works alleviate this problem by fine-tuning the base model~\citep{Zeng2021CertifiedRT,Ye2020SAFERAS} or training an external denoiser~\citep{Salman2020DenoisedSA} to augment the base model with better performances on masked texts.
Despite the effectiveness, these methods require access to the parameters of LLMs, which is often unavailable, and could result in large computational costs.
In the following, we will show how to use LLM itself as a denoiser in a self-denoising manner.

Our objective is to improve the randomized smoothing certification radius on existing LLMs with no access to parameters and no further training.
Specifically, we add an additional denoising step with a denoiser $D(\cdot)$, which processes the masked input before feeding it to the base LLM, \ie,
\vspace{-3mm}
\begin{equation}
\footnotesize
    g'(\vx) = \argmax_{c\in\mathcal{Y}} 
    P_{\vs \sim \mathcal{U}(L, k)}(f(D(\mathcal{M}(\vx, \vs)))=c)\text{.}
    \label{eq:denoised_smoothed_classifier}
\end{equation}
\vspace{-7mm}

\noindent The denoiser is expected to augment the base model to be more robust towards random masks on the inputs.
Specifically, we consider two design choices for the denoiser, 1) instruct the LLM itself to recover the original input $\vx$ given the masked input, and 2) directly remove the masks.
To use the LLM as the denoiser, we use in-context learning to teach the LLM to fill in the masked positions so that the completed sentence is fluent and could preserve the original semantic.
The prompt we used to instruct the LLM could be seen in Appendix~\ref{app:setup}.
On the other hand, we note that when mask rate $m$ is high, such a filling-in-mask may fail to capture the original semantic due to limited remaining words and thus lead to undesired denoising results.
Therefore, under such scenarios, we directly remove the \emph{[Mask]} in the masked positions and use the remaining parts for the next step downstream prediction.

\begin{figure}
    \centering
    \includegraphics[width=0.48\textwidth]{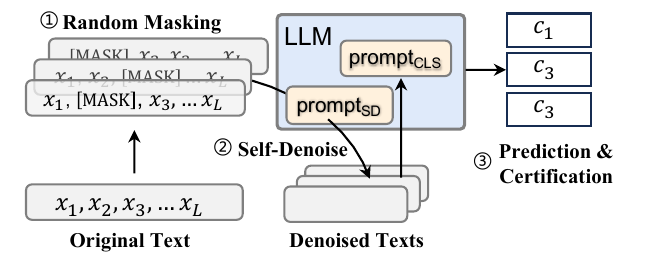}
    \caption{\small Prediction and certification process with our self-denoised smoothed classifier $g(\vx')$.}
    \label{fig:denoiser}
    \vspace{-6mm}
\end{figure}

The prediction and certification pipeline of \ours could be seen in Figure~\ref{fig:denoiser}, where a Monte Carlo algorithm is used for estimating $g'(\vx)$.
The input sentence is firstly perturbed with random masking multiple times.
Different from the original randomized smoothing (with only step \ding{172} and \ding{174} in the figure), we additionally add a denoising step, where the perturbed inputs are fed into the denoiser.
The returned denoised results are fed into the LLM for downstream task prediction, and all predicting results are then integrated to get the final prediction following Equation~\ref{eq:denoised_smoothed_classifier}.
The certification process follows the original randomized smoothing\footnote{
The detailed algorithm could be seen in \citet{zeng2021certified} Algorithm 2, Line 13-24.
} with our smoothed classifier $g'(\vx)$.

\vspace{-2mm}
\section{Experiment}
\label{sec:experiment}
\subsection{Experiment Setup}
\label{sec:experiment_setup}
\paragraph{Dataset and models}
We use the \texttt{SST-2}~\citep{socher-etal-2013-recursive} and \texttt{Agnews}~\citep{zhang2015character} datasets in our experiments. 
We randomly divide the original testing set of \texttt{Agnews} into  two parts equally as the new validation set and testing set and use the official split of the SST-2 dataset.
We use the validation set for model selection and the testing set for evaluation.
We consider Alpaca~\citep{alpaca} as the base LLM to be robustified.
We design prompts with in-context learning to instruct Alpaca to perform the corresponding tasks. 
See details in Appendix~\ref{app:setup}.

\vspace{-2mm}
\paragraph{Evaluation metrics}
Following \citet{zeng2021certified}, we evaluate our methods together with all baselines with both \text{certified accuracy} and empirical robust accuracy. 
The \text{certified accuracy} is calculated for each perturbation scale $d$ over $1\%$ to $10\%$, \ie, $\text{certified accuracy}=\frac{1}{n} \sum_{i=1}^n \mathbb{I}(r_i \geq d)$, where $r_i$ is the certification radius for $i$-th input over in total $n$ examples.
The empirical robust accuracy is calculated using state-of-the-art adversarial attack methods DeepWordBug~\citep{Gao2018BlackBoxGO} and TextBugger~\citep{Li2018TextBuggerGA}.
Specifically, the attackers are adopted to attack the smoothed classifier with at most $10\%$ words perturbation on each sentence, and the accuracy on the attacked adversarial examples are reported.
We also report the clean accuracy on standard examples.

\vspace{-2mm}
\paragraph{Baselines and implementation details}
We compare our method \ours with the randomized smoothing-based certification method \textsc{RanMask} for certified accuracy.
Note that another similar certification method \textsc{Safer} does not consider the same definition for certified robustness with us\footnote{See Section~\ref{sec:related_work} for more explanations.}, so we only compare our method with them on empirical robust accuracy.
The performances of the vanilla base model, termed \textsc{Alpaca}, are also reported.
All baselines are evaluated with the same base model without any finetuning.
The best hyper-parameters of each method are searched on the validation set. 
See details in Appendix~\ref{app:setup}.

\begin{figure}[t]
    \centering
    \resizebox{0.49\textwidth}{!}{    
    \begin{tabular}{ll}
    \hspace{-3mm}
    \includegraphics[width=1.0\textwidth]{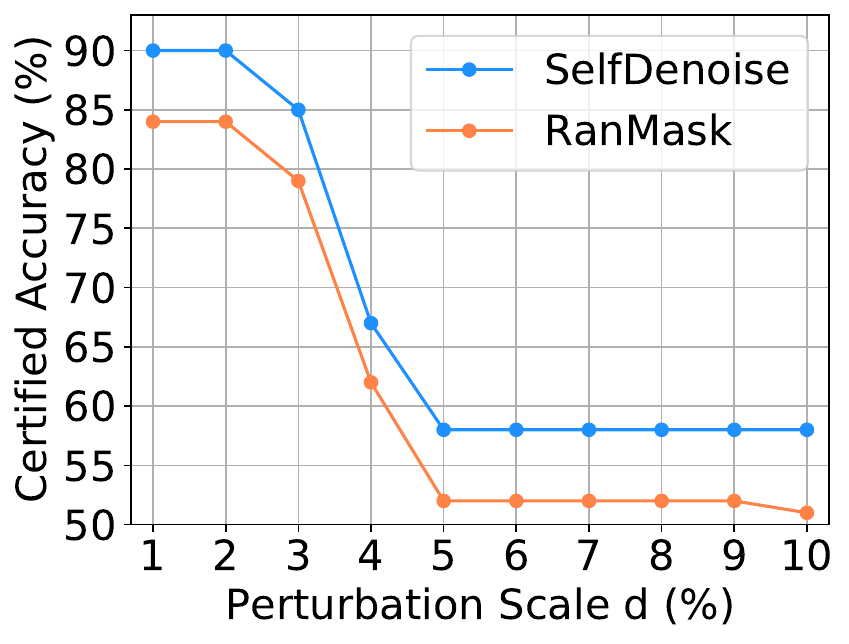} 
         & 
    \hspace{-8mm}
    \includegraphics[width=1.0\textwidth]{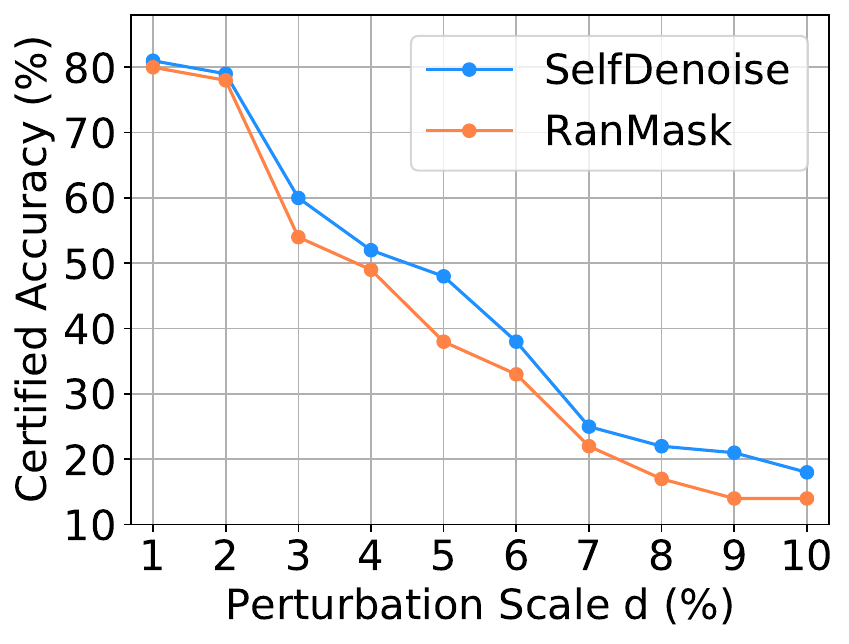}
    \end{tabular}
    }
    \vspace{-3mm}
    \caption{\small Certified accuracy under different perturbation scale $d$ (\%) on \texttt{SST-2} (\emph{left}) and \texttt{Agnews} (\emph{right}).
    }
    \label{fig:certified_ra}
    \vspace{-4mm}
\end{figure}

\begin{table}[t]
    \centering
    \resizebox{0.46\textwidth}{!}{
    \begin{tabular}{c|c|c|c|c}
    \toprule[1pt]
        \midrule
        \multirow{2}{*}{Dataset} & \multirow{2}{*}{Method} & \multirow{2}{*}{\textit{Clean Acc.} (\%)} & \multicolumn{2}{c}{\textit{Empirical Robust Acc.} (\%)} \\
        & & & DeepWordBug & TextBugger \\
        \midrule
        \multirow{4}{*}{\texttt{SST-2}}
        & \textsc{Alpaca} & 89.0 & 52.0 & 45.0\\
        & \textsc{Safer} & 85.0 & 57.0  & 54.0\\
        & \textsc{RanMask} & 84.0  & 52.5 & 48.0 \\
        & \ours & \textbf{90.0} & \textbf{64.5} & \textbf{55.5}\\  
        \midrule
        \multirow{4}{*}{\texttt{Agnews}}
        & \textsc{Alpaca} & \textbf{85.0} & 58.5 & 50.5\\
        & \textsc{Safer} & 83.0 & 55.5 &  53.0 \\
        & \textsc{RanMask} & 82.0 & 58.0 & 53.0 \\
        & \ours & \text{84.0} & \textbf{70.0} & \textbf{66.0}\\ 
        \midrule
    \bottomrule[1pt]
    \end{tabular}}
    \caption{\small Clean accuracy and empirical robust accuracy under DeepWordBug attack and TextBugger attack.}
    \label{tab:empirical_ra}
    \vspace{-4mm}
\end{table}

\vspace{-1mm}
\subsection{Experiment Results}
\label{sec:experiment_results}
Figure~\ref{fig:certified_ra} shows the certification results of the proposed \ours and baseline \textsc{RanMask} on both \texttt{SST-2} and \texttt{Agnews}.
We show that our method could effectively improve certified accuracy beyond \textsc{RanMask} in both two datasets under all perturbation scales.
For example, with $d=5$, our method outperforms \textsc{RanMask} by 11.5\% in \texttt{SST-2} and 26.3\% in \texttt{Agnews}.

We further present the empirical robust accuracy (with at most 10\% word perturbation) of the proposed \ours and baselines in Table~\ref{fig:certified_ra}.
Here are our key observations. \underline{First},
we show our method achieves the best empirical robust accuracy in both two datasets under both attack methods.
Specifically, \ours improves the empirical robust accuracy by 13.2\% in \texttt{SST-2} and 19.7\% in \texttt{Agnews} compared with the second best method under DeepWordBug attack, with 2.8\% and 24.5\% improvements under TextBugger.
\underline{Second}, the proposed method demonstrates a better trade-off between robustness and standard accuracy.
Specifically, our method achieves the best clean accuracy and empirical robust accuracy in \texttt{Agnews}.
In \texttt{SST-2}, \ours introduces 19.7\% improvement in empirical robust accuracy with only a 1.2\% drop in clean accuracy, compared with the vanilla \textsc{Alpaca}.

\vspace{-2mm}
\section{Conclusion}
In this paper, we proposed a randomized smoothing based LLM certification method, \ours, which introduces a self-denoising framework to augment the original LLM by instructing the LLM to act as an additional denoiser, leading to larger certification radius of LLMs.
The proposed could be used as a plug-in module for any LLM without any access to parameters, and no training is needed.
Results from extensive experiments have demonstrated our superiority on both certified robustness and empirical robustness compared with existing works.
For future works, we plan to replace our greedy self-denoising strategy with more plausible choices. 
We will investigate ways to find the optimal strategy by combining vast potential denoising transformations beyond mask filling.

\clearpage
\section{Broader Impacts}
By developing a self-denoising method to enhance the robustness of LLMs in the presence of noisy inputs, this work addresses a key limitation of LLMs and enables their application in high-stake environments. 
The ability to utilize LLMs in these scenarios can have significant positive impacts across various domains, such as healthcare, transportation, and finance, where safety and reliability are critical.
By providing certified guarantees in safety-critical domains, our method can help build more reliable and responsible LLM systems.

Besides, our research contributes to the broader fields of machine learning and artificial intelligence. 
By tackling the challenge of robustness to noisy inputs in LLMs, we advance the understanding and the methodologies in this area. 
This can inspire further research and innovation, leading to improved techniques for enhancing the performance and reliability of LLMs and other machine learning models.

However, it is important to acknowledge the potential biases that may exist in LLMs, as our method relies on them as base models. 
Biases can arise from the training data used for LLMs, and these biases may be propagated by our method. 
We are committed to addressing the issue of biases and promoting fairness and transparency in machine learning systems. 
To mitigate these concerns, we will include proper licenses in the released codes and notify users about the potential risks associated with biases. 
This way, users can be informed and take appropriate measures to address any biases that may arise from the use of our method.
\section{Limitations}
\label{sec:limitations}
Despite the large improvements, our method suffers from the limitation of running time, \ie,  the prediction and certification process is time-consuming.
This is largely because of the $p_c(\vx)$ calculation in Equation~\ref{eq:denoised_smoothed_classifier}.
Such a problem is shared across all randomized smoothing-based methods.
Besides, the additional self-denoising process also brings further computational loads.
It would be interesting to either apply recent works on distributed computation to accelerate our method or develop new large language models specifically for denoising to overcome this issue.

\bibliography{references}

\clearpage
\appendix
\lstset{
    basicstyle=\ttfamily,
    breaklines=true,
    breakindent=0pt,
    frame=single, 
    frameround=tttt 
}
\renewcommand{\lstlistingname}{}
\section{Additional Experiment Setup}
\label{app:setup}
\subsection{Prompts and Instructions}
The prompts and instructions we used for in-context learning on downstream task prediction and self-denoising are shown as follows. 

\begin{lstlisting}[caption={Prompt template used for Alpaca.}]
Below is an instruction that describes a task, paired with an input that provides further context. Write a response that appropriately completes the request.

### Instruction:
{}

### Input:
{}

### Response:
\end{lstlisting}

The following instructions are used to fill in the contents under the ``Instruction'' section. The content under ``Input'' should be filled with different input texts.

\begin{lstlisting}[caption={The instruction used for classification on SST-2.}]
Given an English sentence input, determine its sentiment as positive or negative.
\end{lstlisting}

\begin{lstlisting}[caption={The instruction used for self-denoising on SST-2.}]
Replace each mask word [MASK] in the input sentence with a suitable word. The output sentence should be natural and coherent and should be of the same length as the given sentence.

### Input: 
[MASK] reassembled from [MASK] cutting-room [MASK] of any [MASK] daytime [MASK].

### Response:
apparently reassembled from the cutting-room floor of any given daytime soap.

### Input: 
a [MASK], funny and [MASK] transporting re-imagining [MASK] [MASK] and the beast and 1930s [MASK] films

### Response:
a stirring, funny and finally transporting re-imagining of beauty and the beast and 1930s horror films
\end{lstlisting}

\begin{lstlisting}[caption={The instruction used for classification on Agnews.}]
Given a news article title and description, classify it into one of the four categories: Sports, World, Technology, or Business. Return the category name as the answer.

### Input: 
Title: Venezuelans Vote Early in Referendum on Chavez Rule (Reuters)
Description: Reuters - Venezuelans turned out early and in large numbers on Sunday to vote in a historic referendum that will either remove left-wing President Hugo Chavez from office or give him a new mandate to govern for the next two years.

### Response:
World

### Input:
Title: Phelps, Thorpe Advance in 200 Freestyle (AP)
Description: AP - Michael Phelps took care of qualifying for the Olympic 200-meter freestyle semifinals Sunday, and then found out he had been added to the American team for the evening's 400 freestyle relay final. Phelps' rivals Ian Thorpe and Pieter van den Hoogenband and teammate Klete Keller were faster than the teenager in the 200 free preliminaries.

### Response:
Sports

### Input:
Title: Wall St. Bears Claw Back Into the Black (Reuters)
Description: Reuters - Short-sellers, Wall Street's dwindling band of ultra-cynics, are seeing green again.

### Response:
Business
        
### Input:
Title: 'Madden,' 'ESPN' Football Score in Different Ways (Reuters)
Description: Reuters - Was absenteeism a little high\on Tuesday among the guys at the office? EA Sports would like to think it was because "Madden NFL 2005" came out that day, and some fans of the football simulation are rabid enough to take a sick day to play it.

### Response:
Technology
\end{lstlisting}

\begin{lstlisting}[caption={The instruction used for self-denoising on Agnews.}]
Replace each masked position "[MASK]" in the provided sentence with a suitable word to make it natural and coherent. Only one word should be used to replace each "[MASK]". The returned sentence should be of the same length as the given sentence. Provide the answer directly.
\end{lstlisting}

\subsection{Hyperparameter }
We evaluate on 100 testing instances for certified accuracy in Figure~\ref{fig:certified_ra} and 200 instances for empirical robust accuracy in Table~\ref{tab:empirical_ra}. 
To use the Alpaca for self-denoising, we use beam search for generation and set the repetition penalty to 1.3 and the number of beams to 2. 
We use 500 instances for estimating $\underline{p_c(\vx)}$ with Monte Carlo in the certification process.
In Figure~\ref{fig:certified_ra}, for each perturbation scale, we search the best mask rate $m$ from $\{10\%, 20\%, \ldots, 90\%\}$ on the validation set for our method and \textsc{RanMask}. The best mask rates for each perturbation scale are listed in Table~\ref{tab:best_mask_rate}.
When mask rate $m$ is greater than or equal to $70\%$, we use the removing mask strategy; otherwise, we use Alpaca itself as the denoiser.
For empirical robustness results in Table~\ref{tab:empirical_ra}, we observe that smaller mask rates bring better empirical robust accuracy in the validation set, so we use $m=5\%$ for all methods.

\begin{table}[t]
    \centering
    \resizebox{0.45\textwidth}{!}{
    \begin{tabular}{c|c|cccccccccc}
    \toprule[1pt]
        \midrule
        \multirow{2}{*}{Dataset} & \multirow{2}{*}{Method} & \multicolumn{10}{c}{Perturbation Scale d (\%)} \\
        &  & 1 & 2 & 3 & 4 & 5 & 6 & 7 & 8 & 9 & 10 \\
        \midrule
        \multirow{2}{*}{\texttt{SST-2}} & \textsc{RanMask} & 10 & 10 & 10 & 10 & 80 & 80 & 80 & 80 & 80 & 80 \\
         & \ours & 20 & 20 & 30 & 30 & 70 & 80 & 80 & 90 & 90 & 90 \\
         \midrule
        \multirow{2}{*}{\texttt{Agnews}} & \textsc{RanMask} & 20 & 20 & 70 & 70 & 80 & 80 & 90 & 90 & 90 & 90 \\
        & \ours & 50 & 50 & 70 & 80 & 80 & 80 & 90 & 90 & 90 & 90 \\
        \midrule
        \bottomrule[1pt]
    \end{tabular}}
    \caption{\small The best mask rate $m$ (\%) for each perturbation scale on \texttt{SST-2} and \texttt{Agnews} for \ours and \textsc{RanMask}.}
    \label{tab:best_mask_rate}
\end{table}

\end{document}